\pdfoutput=1

\documentclass[11pt]{article}

\usepackage[]{acl}

\usepackage{times}
\usepackage{latexsym}
\usepackage{graphicx}
\usepackage{booktabs}
\usepackage{amsmath}
\usepackage{multirow} 
\usepackage{xspace}
\usepackage{enumitem}
\usepackage{multicol}
\usepackage{float}
\usepackage{subcaption,amsfonts,dcolumn}
\usepackage{xparse}
\usepackage{fancyvrb}
\NewDocumentCommand{\pritika}{ mO{} }{\textcolor{purple}{\textsuperscript{\textit{pritika}}\textsf{\textbf{\small[#1]}}}}

\newcommand{\datasetname}{\textsc{{DescToTTo}}\xspace}
\newcommand{\eval}{\textsc{{TabEval}}\xspace}

\usepackage[T1]{fontenc}

\usepackage[utf8]{inputenc}

\usepackage{microtype}

\usepackage{inconsolata}

\title{Is This a Bad Table? \\ A Closer Look at the Evaluation of Table Generation from Text}

\author{Pritika Ramu \quad Aparna Garimella \quad Sambaran Bandyopadhyay \\
  Adobe Research, India\\
  \texttt{\{pramu,garimell,sambaranb\}@adobe.com}\\}

\begin{document}
\maketitle

\begin{abstract}
Understanding whether a generated table is of good quality is important to be able to use it in creating or editing documents using automatic methods. 
In this work, we underline that existing measures for table quality evaluation fail to capture the overall semantics of the tables, and sometimes unfairly penalize good tables and reward bad ones.
We propose \textbf{\eval}, a novel table evaluation strategy that captures table semantics by first breaking down a table into a list of natural language atomic statements and then compares them with ground truth statements using entailment-based measures.
To validate our approach, we curate a dataset comprising of text descriptions for 1,250 diverse Wikipedia tables, covering a range of topics and structures, in contrast to the limited scope of existing datasets.
We compare \eval with existing metrics using unsupervised and supervised text-to-table generation methods, demonstrating its stronger correlation with human judgments of table quality across four datasets.
\end{abstract}

\section{Introduction}

Tables are an integral form of representing content in real-world documents such as news articles, financial reports, and contracts. Document generation requires the generation of high-quality tables along with other modalities.
While the problems of table-to-text generation and table summarization have been widely studied \cite{parikh-etal-2020-totto,chen2022towards,guo2023towards}, text-to-table generation has been gaining increasing attention more recently \cite{wu-etal-2022-text-table,li-etal-2023-sequence-sequence}.

Differentiating between good and bad quality tables generated from text is crucial for their usability in documents. Failure to accurately assess table quality can result in including subpar content or overlooking valuable tables in documents. 

\begin{figure}[t!]
    \centering
    \includegraphics[width=0.9\columnwidth]{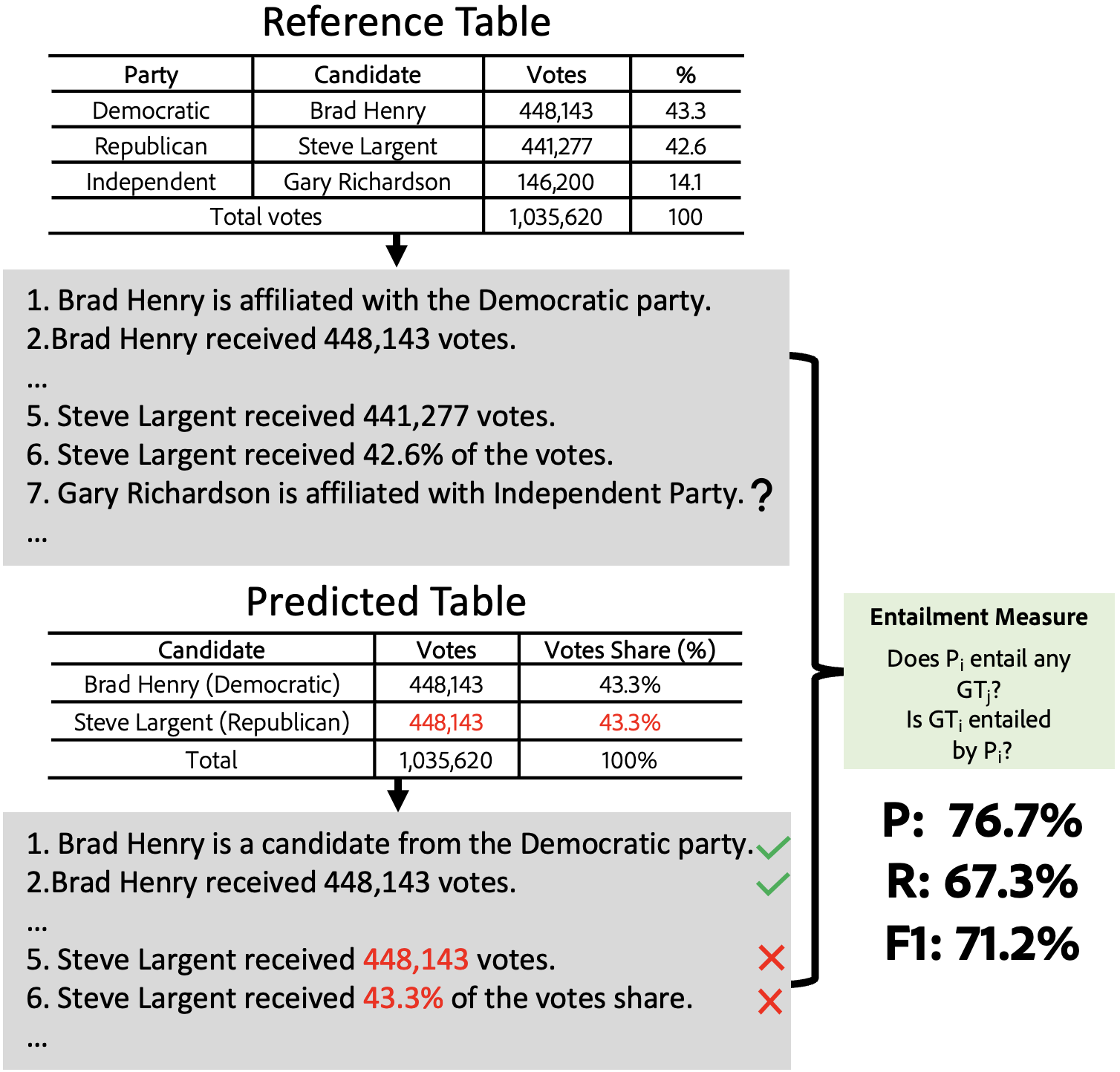}
    \caption{\footnotesize Tables are unrolled using TalUnroll prompting with an LLM, and the obtained statements are evaluated using NLI.}
    \label{fig:intro}
    \vspace{-0.15in}
\end{figure}

Existing text-to-table works adopt metrics based on exact match and BertScore \cite{bert-score} of the header cells of generated tables with the ground truth ones, and for the non-header cells, they use the header cell information also to compare the resulting tuples.
However, a major limitation with such measures is that they evaluate the table cells (or tuples) independently without considering contextual information from the neighboring cells.
This can lead to incorrect penalization of good tables, or incorrect rewarding of bad tables. 


In this paper, we {\bf first} propose {\bf \eval}, a two-staged table evaluation approach that views tables holistically rather than considering values independently while evaluating their quality.
Given the table intent, reference, and predicted table, we first {\it unroll} the tables into sets of meaningful natural language (NL) statements that convey the overall table semantics.
We propose {\sc TabUnroll}, a novel prompting technique to unroll a table using Chain-of-Thought \cite{kojima2023large,wei2023chainofthought} using an LLM.
We then compute the entailment scores between the unrolled NL statements of predicted and ground truth tables and provide an aggregate as the measure of table quality.

Existing datasets used for text-to-table generation, such as Rotowire \cite{wiseman-etal-2017-challenges}, Wikibio \cite{lebret-etal-2016-neural}, WikiTableText \cite{bao2018tabletotext}, are restricted in domain and schema.
Our {\bf second} contribution is curation of a dataset consisting of 1,250 general domain tables along with their textual descriptions, to assess our evaluation strategy across different domains.

{\bf Thirdly,} we perform several experiments utilizing existing text-to-table methods and LLM-based prompting techniques. 
We collect human ratings for table quality on test generations obtained using from various method-dataset combinations. 
\eval shows significantly higher correlations with human ratings compared to the existing metrics across most scenarios. 
We highlight important failure cases of the existing metrics qualitatively, while underlining limitations of ours too to facilitate further research on evaluating the quality of automatic table generation methods in documents.

\section{Proposed Evaluation Strategy}
We introduce \textbf{\eval}, a two-stage pipeline (Fig. \ref{fig:intro}) that evaluates the semantic quality of generated tables against a reference table to ensure they convey the same information.
\paragraph{Table Unrolling.}
We propose \textbf{TabUnroll}, a prompting strategy using Chain-of-Thought to unroll a table into meaningful NL atomic statements. 
The input is the table intent (table name/ caption/ description) and the table in HTML. 
It follows a generalizable schema outlined in \cite{wang-etal-2022-super}---{\bf (1) Instruction set:} LLM is prompted to identify the column headers, rows, and suitable column(s) serving as primary key(s) to depict each unit of information conveyed by the table.
We define the primary key as the column(s) that contains values that uniquely identify each row in a table. 
We provide instructions to use the identified primary key(s) as anchor(s) to construct meaningful atomic statements by using values from the rest of the columns one at a time.
In the absence of primary key, we instruct to form the statements by picking as few columns (two or above) as possible to form meaningful statements.
The LLM is also prompted to attribute the specific rows from which the atomics are constructed in the form on inline citations, to mitigate any hallucinations \cite{wei2023chainofthought}. 
{\bf (2) Few-shot examples:} We provide positive and negative examples of how tables should be unrolled. Given that LLMs tend to struggle with negation tasks \cite{truong-etal-2023-language}, we show examples of what not to produce. (Appendix \ref{sec:unrolled_tables} has the full prompt template and sample unrolled tables.)

\paragraph{Entailment-based Scoring.} 

After obtaining the unrolled statements from the ground truth and predicted tables (of sizes $M$ and $N$ respectively), we employ Natural Language Inference \cite{liu2019roberta} to determine whether the information conveyed by the predicted table is also present in the ground truth table, and vice versa. 

Precision (Correctness) is computed as the average of the maximum entailment scores between each predicted statement \( p_i \) and all ground truth statements \( g_j \), Recall (Completeness) as the average of the maximum entailment scores between each ground truth statement \( g_j \) and all predicted statements \( p_i \) and F1 (Overall quality) as the harmonic mean of precision and recall.

{\scriptsize
\begin{equation}
\text{Precision} = \frac{\sum_{i=1}^{N} \max_{j=1}^{M} \text{score}(p_i, g_j)}{N}
\end{equation}
\begin{equation}
\text{Recall} = \frac{\sum_{j=1}^{M} \max_{i=1}^{N} \text{score}(p_i, g_j)}{M}
\end{equation}
}


\begin{table}[t]
  \centering
  \small
  \scalebox{0.74}{
    \begin{tabular}{l|r|r|r|r}
      \toprule
      \textbf{Statistic} & \textbf{DescToTTo} & \textbf{RotoWire} & \textbf{WikiBio} & \textbf{WikiTableText} \\ 
      \midrule
      \# tables (train) & 1,000 & 3.4k & 3.4k & 10k \\ 
      \# tables (test) & 250 & 728 & 728 & 1.3k \\ 
      Avg. text length & 155.94 & 351.05 & 122.3 & 19.59 \\ 
      Avg. \# rows & 5.66 & 2.71/7.26 & 4.2 & 4.1 \\ 
      Avg. \# cols  & 5.43 & 4.84/8.75 & 2 & 2 \\ 
      Multirow/ col & Yes & No & No & No \\
      \# multirow/ col & 276 & - & - & - \\
      tables & & & & \\
      Domain & Wikipedia & Sports & Bio & Wikipedia \\ 
      \bottomrule
    \end{tabular}
  }
\caption{Comparative statistics of the datasets.}
\label{tab:dataset}
\vspace{-0.15in}
\end{table}

\begin{table*}[ht]
  \centering
  \small
  \scalebox{0.65}{
  \begin{tabular}{l|c|c|c|c|c|c|c|c|c|c|c|c|c|c|c|c|c|c|c|c|c}
    \toprule
    \multicolumn{2}{c|}{} & \multicolumn{5}{c|}{\datasetname} & \multicolumn{5}{c|}{\sc RotoWire} & \multicolumn{5}{c|}{\sc Wikibio} & \multicolumn{5}{c}{\sc WikiTableText} \\
    \midrule
    Metric & Model & E & Chrf & BS & O-C & O-G & E & Chrf & BS & O-C & O-G & E & Chrf & BS & O-C & O-G & E & Chrf & BS & O-C & O-G \\ 
    \midrule
    \multirow{4}{*}{Corct.} & GPT-4 & 0.09 & 0.10 & 0.21 & \textbf{0.35} & 0.33 & 0.12 & 0.14 & 0.36 & \textbf{0.45} & 0.44 & 0.18 & 0.23 & 0.57 & \textbf{0.61} & 0.60 & 0.19 & 0.28 & 0.57 & \textbf{0.59} & \textbf{0.59}\\
    & GPT-3.5 & 0.09 & 0.11 & 0.22 & \textbf{0.36} & 0.33 & 0.13 & 0.16 & 0.36 & \textbf{0.44} & \textbf{0.44} & 0.18 & 0.23 & 0.57 & \textbf{0.60} & \textbf{0.60} & 0.19 & 0.28 & 0.56 & \textbf{0.58} & \textbf{0.58}\\
    & L-IFT & 0.11 & 0.18 & 0.27 & \textbf{0.39} & 0.36 & 0.26 & 0.27 & 0.38 & \textbf{0.48} & \textbf{0.48} & 0.30 & 0.39 & \textbf{0.63} & 0.62 & 0.62 & 0.31 & 0.42 & 0.60 & \textbf{0.61} & \textbf{0.61} \\
    & Seq2Seq & 0.15 & 0.20 & 0.31 & \textbf{0.41} & 0.37 & 0.30 & 0.34 & 0.37 & \textbf{0.51} & 0.50 & 0.32 & 0.42 & \textbf{0.64} & 0.62 & 0.62 & 0.32 & 0.43 & \textbf{0.63} & \textbf{0.63} & 0.62\\
    \midrule
    \multirow{4}{*}{Compl.} & GPT-4 & 0.08 & 0.11 & 0.37 & \textbf{0.41} & 0.39 & 0.08 & 0.12 & 0.37 & \textbf{0.46} & 0.45 & 0.19 & 0.27 & 0.59 & \textbf{0.64} & \textbf{0.64} & 0.19 & 0.26 & 0.59 & \textbf{0.62} & \textbf{0.62}\\
    & GPT-3.5 & 0.07 & 0.14 & 0.35 & \textbf{0.40} & 0.38 & 0.09 & 0.13 & 0.39 & \textbf{0.44} & \textbf{0.44} & 0.18 & 0.26 & 0.57 & \textbf{0.62} & 0.61 & 0.17 & 0.25 & 0.56 & \textbf{0.61} & 0.60\\
    & L-IFT & 0.28 & 0.32 & 0.40 & \textbf{0.45} & 0.42 & 0.31 & 0.35 & 0.43 & \textbf{0.47} & 0.46 & 0.35 & 0.40 & 0.63 & \textbf{0.64} & \textbf{0.64} & 0.34 & 0.38 & \textbf{0.65} & \textbf{0.65} & \textbf{0.65}\\
    & Seq2Seq & 0.29 & 0.32 & 0.43 & \textbf{0.46} & 0.42 & 0.32 & 0.35 & 0.43 & \textbf{0.48} & 0.47 & 0.36 & 0.42 & \textbf{0.66} & \textbf{0.66} & 0.65 & 0.34 & 0.40 & \textbf{0.64} & 0.63 & 0.63\\
    \midrule
    \multirow{4}{*}{Ovrl.} & GPT-4 & 0.07 & 0.10 & 0.12 & \textbf{0.37} & 0.36 & 0.07 & 0.09 & 0.30 & \textbf{0.42} & 0.41 & 0.18 & 0.24 & 0.58 & \textbf{0.62} & 0.61 & 0.19 & 0.27 & 0.58 & \textbf{0.61} & 0.60\\
    & GPT-3.5 & 0.07 & 0.11 & 0.12 & \textbf{0.37} & 0.36 & 0.06 & 0.10 & 0.26 & \textbf{0.41} & 0.40 & 0.18 & 0.24 & 0.57 & \textbf{0.61} & \textbf{0.61} & 0.18 & 0.26 & 0.56 & \textbf{0.59} & \textbf{0.59}\\
    & L-IFT & 0.15 & 0.19 & 0.24 & \textbf{0.36} & 0.35 & 0.28 & 0.31 & 0.36 & \textbf{0.39} & 0.37 & 0.32 & 0.39 & \textbf{0.63} & \textbf{0.63} & \textbf{0.63} & 0.32 & 0.39 & \textbf{0.63} & \textbf{0.63} & 0.62\\
    & Seq2Seq & 0.14 & 0.17 & 0.21 & \textbf{0.34} & \textbf{0.34} & 0.26 & 0.30 & 0.34 & \textbf{0.37} & 0.36 & 0.34 & 0.41 & \textbf{0.65} & {0.64} & 0.64 & 0.33 & 0.41 & \textbf{0.63} & \textbf{0.63} & \textbf{0.63}\\
    \bottomrule
  \end{tabular}
  }
  \caption{\footnotesize The correlations of our metric and existing ones with human ratings. Corct: Correctness, Compl: Completeness, Ovrl: Overall, L-IFT: LLaMa-2 IFT; O-C: Our metric with Claude-based unrolling; O-G: Our metric with GPT-4 unrolling.}
  \label{tab:results}
  \vspace{-0.15in}
\end{table*}

\section{Dataset Curation}
Table-to-text datasets, like Wikibio \cite{lebret-etal-2016-neural}, WikiTableText \cite{bao2018tabletotext}, and E2E \cite{novikova-etal-2017-e2e}, contain simple key-value pairs for tables. Rotowire \cite{wiseman-etal-2017-challenges} offers more complex tables, but specific to sports domain with fixed schema, with columns and rows for player/team statistics and names respectively.
\textsc{ToTTo} dataset \cite{parikh-etal-2020-totto} offers a diverse range of Wikipedia tables from different domains and schemas, providing a broad representation of tables found in documents. 
However, its annotations are tailored for creating text descriptions of individual rows, not whole tables, making it unsuitable for generating tables from these descriptions.

To have a general-domain text-to-table evaluation, we curate \textbf{\datasetname}, by augmenting tables from {\sc ToTTo} with parallel text descriptions. 
It comprises of 1,250 tables, each annotated with \textit{table text} and \textit{intent}.
Annotators, fluent in English and skilled in content writing, are recruited from a freelancing platform and compensated at $\$15/hour$. 
They are selected based on a pilot test where six candidates are to annotate five samples each. 
The outputs are rated by two judges; 3 annotators are approved by them. 
They are instructed to provide parallel descriptions (\textit{table text}) and intents for tables, using Wikipedia article for context. 
Each table is annotated by one of the three annotators.
Samples validated by judges are included in the final set.
They belong to diverse topics including sports, politics, entertainment, arts, and so on.
They include hierarchical tables with multiple rows and/ or columns, thus adding to their schema-wise diversity (Table \ref{tab:dataset}).
The table texts contain $6.53$ sentences on average, and the tables are of varied sizes ranging from 1x1 upto 18x33 dimensions (examples in Appendix \ref{sec:dataset}).

\section{Experiments}
To validate \eval, we conduct experiments using four text-to-table generation models on four datasets. 
In the supervised setting, we perform instruction fine-tuning on \texttt{llama-2-7b-chat-hf}, and use the Seq2Seq text-to-table baseline proposed by \citet{wu-etal-2022-text-table}.
Tables generated by \texttt{gpt-4} and \texttt{gpt-3.5-turbo} models are in an unsupervised setting with few-shot examples. 
NVIDIA A100 GPUs were used for LLaMa IFT. The prompts for GPT and LLaMa IFT are in Appendix \ref{sec:table_prompt}. 
In \eval, we experiment with \texttt{gpt-4} and \texttt{Claude-3-Opus} \cite{anthropic2024claude} for table unrolling, and use \texttt{roberta-large-mnli} \cite{liu2019roberta} for measuring entailment.\\
\noindent
\textbf{Baselines.}
We compare \eval with those in \cite{wu-etal-2022-text-table}, which assess tables by representing them as tuples (row header, cell value)/ triples (row header, col header, cell value) and comparing them with ground truth tuples/ triples for exact matches (E), chrf \cite{popovic-2015-chrf}, and rescaled BertScore (BS) \cite{bert-score}. \\
\noindent
{\bf Metrics.} We obtain human ratings (1-5 scale) for correctness, completeness, and overall quality of generated tables, comparing them to reference (instructions in Appendix \ref{sec:human}). We calculate the Pearson correlation between our metric scores and human ratings, comparing these to baseline metrics.


\begin{table}[ht]
\centering
\scriptsize
\setlength{\tabcolsep}{3pt} 
\scalebox{0.87}{
\begin{tabular}{@{}lcccccccccc@{}}
\toprule
        & \multicolumn{5}{c}{DescToTTo}                       & \multicolumn{5}{c}{Rotowire}                       \\
        \cmidrule(lr){2-6} \cmidrule(lr){7-11}
Model   & E & Chrf & BS & O-C   & O-G  & E & Chrf & BS & O-C   & O-G \\
\midrule
GPT-4   & 35.27 & 37.43 & 41.78 & 67.96 & 68.92 & 56.28 & 58.15 & 63.99 & 77.63 & 77.54 \\
GPT-3.5 & 34.14 & 37.68 & 40.99 & 65.82 & 67.14 & 33.27 & 35.96 & 57.89 & 77.09 & 77.15 \\
L-IFT   & 47.13 & 49.44 & 63.01 & 55.89 & 55.91 & 80.71 & 82.35 & 87.62 & 78.43 & 78.20 \\
Seq2Seq & 34.87 & 37.45 & 46.24 & 46.17 & 50.99 & 82.93 & 84.75 & 89.77 & 80.13 & 81.02 \\
\bottomrule
\end{tabular}}
\caption{\footnotesize Comparison of model performances using various metrics; O-C: Ours with Claude; O-G: Ours with GPT-4.}
\label{tab:abs}
\vspace{-0.15in}
\end{table}

\begin{figure*}[t]
  \centering
  \includegraphics[width=0.85\textwidth]{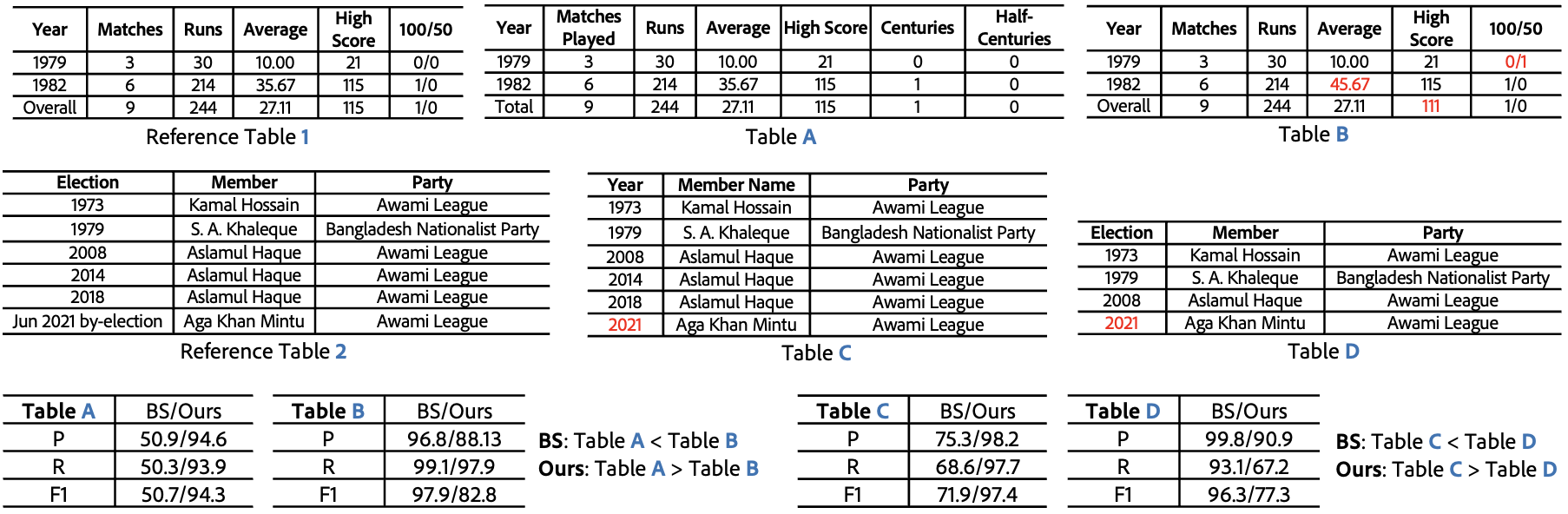}
    \caption{\footnotesize Sample generated tables with precision (P), recall (R), and F1 using \eval with GPT-4 and BertScore-based (BS). BS penalises tables for variation in column headers. Table A, despite having correct details, scores lower with BS but high with ours. Table B, with errors, is appropriately penalized by \eval. Table C covers all the details from reference table, receives lower precision and recall with BS but high scores with ours. Table D, missing some rows, has reduced recall with \eval.}
  \label{fig:examples}
  \vspace{-0.15in}
\end{figure*}
\section{Results \& Discussion}
We obtain human ratings for 1,000 test tables (250 per dataset) from three annotators, with medium to high agreement ($\alpha$: $0.55$, $0.60$, $0.62$ for quality, correctness, and completeness, respectively) \cite{Krippendorff1970}. Pearson correlations are computed between the automatic metrics with these ratings across various dataset-method pairs (Table \ref{tab:results}).
We obtain correlations between metric precision and correctness (human-rated), recall and completeness, and F1 score and overall quality and usability.

\eval has higher correlations than that of the existing metrics across most configurations, indicating that our metric is able to evaluate table semantics more accurately compared to the existing ones.
The increments are higher for \datasetname and RotoWire than for the other two datasets; this is because,
WikiBio and WikiTableText, contain simple key-value pairs that are mostly extractive in nature, and are thus effectively evaluated using the BS-based metric for (row, value) tuples in generated tables, yielding correlation scores comparable to \eval.
Particularly in supervised settings, the correlations are slightly higher using BS on these datasets, as they tend to generate very well-rehearsed generations based on the training data.
RotoWire has a fixed schema for player/team statistics and names, resulting in less structural and terminological variability in its tables compared to \datasetname, which lacks a fixed schema and features more diverse, multirow, and multicolumn table structures.
Thus, the improvements in the correlations of \eval are higher on \datasetname compared to those in RotoWire.
\\
\noindent
{\bf Correctness vs. Completeness.} 
On \datasetname and RotoWire, \eval's correlation improvement over BS is higher for correctness ($+0.11$ avg.) than completeness ($+0.05$ avg.). We observe that missing values in model-generated tables usually occur at the row level, rather than individual values within rows, making BS's individual triple-based recall closer to that of \eval. However, the difference in correlation is starker in the case of correctness, as bad tables with some incorrect values are also rated highly by BS, as the overall table and row semantics are not accounted for by the existing metric, whereas ours accounts for this correctly to a greater degree. Fig. \ref{fig:examples} illustrates this: Table B and D, despite having incorrect values, scores nearly $100\%$ in BS's precision, recall, and F1, while our metric accurately penalizes it.
\\
\noindent
{\bf Unsupervised vs. Supervised settings.}
For \datasetname, unsupervised settings gain higher correlation scores ($+0.25$ avg.) than supervised settings ($+0.13$ avg.). Similarly on RotoWire, unsupervised settings gain more ($+0.13$ avg.) compared to supervised settings ($+0.03$ avg.). In supervised settings, models tend to learn and use specific words and patterns prevalent in the reference tables, adhering closely to the training data. In contrast, LLMs, leveraging their extensive general knowledge, tend to deviate from these specific patterns without fine-tuning though generating semantically accurate tables.
Our metric captures this, as can be seen in the better correlations, particularly in unsupervised or low-supervision scenarios (also seen in Fig. \ref{fig:examples}).

Table \ref{tab:abs} shows the performance of each model using different metrics.
\eval diverges more from existing metrics on \datasetname, which requires deep semantic understanding, than on RotoWire, which involves mainly numerical data.
The existing metrics provide significantly lower scores for GPT-4 than the others on \datasetname (though these generations are often accurate semantically), which would be misleading for users looking for right models for the table generation task; \eval captures this better.

\noindent
{\bf Quality of Unrolling.} To assess the quality of extracted statements, which impacts the final metric quality, we conduct a study to rate the correctness and coverage of statements obtained using GPT-4. 3 annotators of similar backgrounds (post-undergraduate, proficient in English) evaluated 120 tables with their intents and statements, rating each on a 1-5 scale. 
Each table has an average of 15 statements.
The average scores are correctness: $4.67$ and coverage: $4.87$ ($\alpha=0.87$ and $0.87$ respectively).
Further, two annotators are instructed to rate the statements as atomic or not, and meaningful or not: $97.3\%$ statements are rated as atomic by both ({\it i.e.,} can not be broken further down into meaningful statements), and all of them are rated as meaningful.
See Appendix \ref{sec:valid} for task samples.

In this work, we focused on the evaluation of general and domain-specific tables with relatively simpler structures. Future work includes evaluation of more complex tables (e.g., large, nested, or multiple tables from single texts), and evaluating table structures based on their readability in addition to semantics. We also aim to develop a reference-free metric based on \eval, comparing unrolled statements directly against the input text.



\section{Limitations}
Since we rely on LLMs to break down a given table into atomic statements, our method will be limited by the quality of the LLM outputs and any potential hallucinations.
However, we use GPT-4 in our evaluation pipeline, and note that the unrolled statements rarely contain hallucinations.
There is a trade-off while using such large models---while the quality of unrolled statements will be very good, they can be computationally expensive. 
With GPT-3.5 and LLaMa variants, we noted more hallucinations in our preliminary explorations.

In this work, we only focus on the semantic quality of tables; we do not evaluation the structural quality, e.g., understanding the right structure for conveying a given intent in an easy-to-read and visually appealing manner.
This can also form one of the future works for this study.

\bibliography{anthology}

\begin{thebibliography}{18}
\expandafter\ifx\csname natexlab\endcsname\relax\def\natexlab#1{#1}\fi

\bibitem[{Anthropic(2024)}]{anthropic2024claude}
AI~Anthropic. 2024.
\newblock The claude 3 model family: Opus, sonnet, haiku.
\newblock \emph{Claude-3 Model Card}.

\bibitem[{Bao et~al.(2018)Bao, Tang, Duan, Yan, Lv, Zhou, and Zhao}]{bao2018tabletotext}
Junwei Bao, Duyu Tang, Nan Duan, Zhao Yan, Yuanhua Lv, Ming Zhou, and Tiejun Zhao. 2018.
\newblock \href {http://arxiv.org/abs/1805.11234} {Table-to-text: Describing table region with natural language}.

\bibitem[{Chen et~al.(2022)Chen, Lu, Xu, Li, Zhou, Dou, and Xiong}]{chen2022towards}
Miao Chen, Xinjiang Lu, Tong Xu, Yanyan Li, Jingbo Zhou, Dejing Dou, and Hui Xiong. 2022.
\newblock Towards table-to-text generation with pretrained language model: A table structure understanding and text deliberating approach.
\newblock In \emph{The 2022 Conference on Empirical Methods in Natural Language Processing (EMNLP' 22)}.

\bibitem[{Guo et~al.(2023)Guo, Zhou, Qi, Yan, He, Zheng, Lin, Wang, and Zhou}]{guo2023towards}
Zhixin Guo, Jianping Zhou, Jiexing Qi, Mingxuan Yan, Ziwei He, Guanjie Zheng, Zhouhan Lin, Xinbing Wang, and Chenghu Zhou. 2023.
\newblock Towards controlled table-to-text generation with scientific reasoning.
\newblock \emph{arXiv preprint arXiv:2312.05402}.

\bibitem[{Kojima et~al.(2023)Kojima, Gu, Reid, Matsuo, and Iwasawa}]{kojima2023large}
Takeshi Kojima, Shixiang~Shane Gu, Machel Reid, Yutaka Matsuo, and Yusuke Iwasawa. 2023.
\newblock \href {http://arxiv.org/abs/2205.11916} {Large language models are zero-shot reasoners}.

\bibitem[{Krippendorff(1970)}]{Krippendorff1970}
Klaus Krippendorff. 1970.
\newblock Estimating the reliability, systematic error and random error of interval data.
\newblock \emph{Educational and Psychological Measurement}, 30(1):61--70.

\bibitem[{Lebret et~al.(2016)Lebret, Grangier, and Auli}]{lebret-etal-2016-neural}
R{\'e}mi Lebret, David Grangier, and Michael Auli. 2016.
\newblock \href {https://doi.org/10.18653/v1/D16-1128} {Neural text generation from structured data with application to the biography domain}.
\newblock In \emph{Proceedings of the 2016 Conference on Empirical Methods in Natural Language Processing}, pages 1203--1213, Austin, Texas. Association for Computational Linguistics.

\bibitem[{Li et~al.(2023)Li, Wang, Shao, Zheng, Wang, and Su}]{li-etal-2023-sequence-sequence}
Tong Li, Zhihao Wang, Liangying Shao, Xuling Zheng, Xiaoli Wang, and Jinsong Su. 2023.
\newblock \href {https://doi.org/10.18653/v1/2023.findings-acl.330} {A sequence-to-sequence{\&}set model for text-to-table generation}.
\newblock In \emph{Findings of the Association for Computational Linguistics: ACL 2023}, pages 5358--5370, Toronto, Canada. Association for Computational Linguistics.

\bibitem[{Liu et~al.(2019)Liu, Ott, Goyal, Du, Joshi, Chen, Levy, Lewis, Zettlemoyer, and Stoyanov}]{liu2019roberta}
Yinhan Liu, Myle Ott, Naman Goyal, Jingfei Du, Mandar Joshi, Danqi Chen, Omer Levy, Mike Lewis, Luke Zettlemoyer, and Veselin Stoyanov. 2019.
\newblock Roberta: A robustly optimized bert pretraining approach.
\newblock \emph{arXiv preprint arXiv:1907.11692}.

\bibitem[{Novikova et~al.(2017)Novikova, Du{\v{s}}ek, and Rieser}]{novikova-etal-2017-e2e}
Jekaterina Novikova, Ond{\v{r}}ej Du{\v{s}}ek, and Verena Rieser. 2017.
\newblock \href {https://doi.org/10.18653/v1/W17-5525} {The {E}2{E} dataset: New challenges for end-to-end generation}.
\newblock In \emph{Proceedings of the 18th Annual {SIG}dial Meeting on Discourse and Dialogue}, pages 201--206, Saarbr{\"u}cken, Germany. Association for Computational Linguistics.

\bibitem[{Parikh et~al.(2020)Parikh, Wang, Gehrmann, Faruqui, Dhingra, Yang, and Das}]{parikh-etal-2020-totto}
Ankur Parikh, Xuezhi Wang, Sebastian Gehrmann, Manaal Faruqui, Bhuwan Dhingra, Diyi Yang, and Dipanjan Das. 2020.
\newblock \href {https://doi.org/10.18653/v1/2020.emnlp-main.89} {{ToTTo}: A controlled table-to-text generation dataset}.
\newblock In \emph{Proceedings of the 2020 Conference on Empirical Methods in Natural Language Processing (EMNLP)}, pages 1173--1186, Online. Association for Computational Linguistics.

\bibitem[{Popovi{\'c}(2015)}]{popovic-2015-chrf}
Maja Popovi{\'c}. 2015.
\newblock \href {https://doi.org/10.18653/v1/W15-3049} {chr{F}: character n-gram {F}-score for automatic {MT} evaluation}.
\newblock In \emph{Proceedings of the Tenth Workshop on Statistical Machine Translation}, pages 392--395, Lisbon, Portugal. Association for Computational Linguistics.

\bibitem[{Truong et~al.(2023)Truong, Baldwin, Verspoor, and Cohn}]{truong-etal-2023-language}
Thinh~Hung Truong, Timothy Baldwin, Karin Verspoor, and Trevor Cohn. 2023.
\newblock \href {https://doi.org/10.18653/v1/2023.starsem-1.10} {Language models are not naysayers: an analysis of language models on negation benchmarks}.
\newblock In \emph{Proceedings of the 12th Joint Conference on Lexical and Computational Semantics (*SEM 2023)}, pages 101--114, Toronto, Canada. Association for Computational Linguistics.

\bibitem[{Wang et~al.(2022)Wang, Mishra, Alipoormolabashi, Kordi, Mirzaei, Naik, Ashok, Dhanasekaran, Arunkumar, Stap, Pathak, Karamanolakis, Lai, Purohit, Mondal, Anderson, Kuznia, Doshi, Pal, Patel, Moradshahi, Parmar, Purohit, Varshney, Kaza, Verma, Puri, Karia, Doshi, Sampat, Mishra, Reddy~A, Patro, Dixit, and Shen}]{wang-etal-2022-super}
Yizhong Wang, Swaroop Mishra, Pegah Alipoormolabashi, Yeganeh Kordi, Amirreza Mirzaei, Atharva Naik, Arjun Ashok, Arut~Selvan Dhanasekaran, Anjana Arunkumar, David Stap, Eshaan Pathak, Giannis Karamanolakis, Haizhi Lai, Ishan Purohit, Ishani Mondal, Jacob Anderson, Kirby Kuznia, Krima Doshi, Kuntal~Kumar Pal, Maitreya Patel, Mehrad Moradshahi, Mihir Parmar, Mirali Purohit, Neeraj Varshney, Phani~Rohitha Kaza, Pulkit Verma, Ravsehaj~Singh Puri, Rushang Karia, Savan Doshi, Shailaja~Keyur Sampat, Siddhartha Mishra, Sujan Reddy~A, Sumanta Patro, Tanay Dixit, and Xudong Shen. 2022.
\newblock \href {https://doi.org/10.18653/v1/2022.emnlp-main.340} {Super-{N}atural{I}nstructions: Generalization via declarative instructions on 1600+ {NLP} tasks}.
\newblock In \emph{Proceedings of the 2022 Conference on Empirical Methods in Natural Language Processing}, pages 5085--5109, Abu Dhabi, United Arab Emirates. Association for Computational Linguistics.

\bibitem[{Wei et~al.(2023)Wei, Wang, Schuurmans, Bosma, Ichter, Xia, Chi, Le, and Zhou}]{wei2023chainofthought}
Jason Wei, Xuezhi Wang, Dale Schuurmans, Maarten Bosma, Brian Ichter, Fei Xia, Ed~Chi, Quoc Le, and Denny Zhou. 2023.
\newblock \href {http://arxiv.org/abs/2201.11903} {Chain-of-thought prompting elicits reasoning in large language models}.

\bibitem[{Wiseman et~al.(2017)Wiseman, Shieber, and Rush}]{wiseman-etal-2017-challenges}
Sam Wiseman, Stuart Shieber, and Alexander Rush. 2017.
\newblock \href {https://doi.org/10.18653/v1/D17-1239} {Challenges in data-to-document generation}.
\newblock In \emph{Proceedings of the 2017 Conference on Empirical Methods in Natural Language Processing}, pages 2253--2263, Copenhagen, Denmark. Association for Computational Linguistics.

\bibitem[{Wu et~al.(2022)Wu, Zhang, and Li}]{wu-etal-2022-text-table}
Xueqing Wu, Jiacheng Zhang, and Hang Li. 2022.
\newblock \href {https://doi.org/10.18653/v1/2022.acl-long.180} {Text-to-table: A new way of information extraction}.
\newblock In \emph{Proceedings of the 60th Annual Meeting of the Association for Computational Linguistics (Volume 1: Long Papers)}, pages 2518--2533, Dublin, Ireland. Association for Computational Linguistics.

\bibitem[{Zhang* et~al.(2020)Zhang*, Kishore*, Wu*, Weinberger, and Artzi}]{bert-score}
Tianyi Zhang*, Varsha Kishore*, Felix Wu*, Kilian~Q. Weinberger, and Yoav Artzi. 2020.
\newblock \href {https://openreview.net/forum?id=SkeHuCVFDr} {Bertscore: Evaluating text generation with bert}.
\newblock In \emph{International Conference on Learning Representations}.

\end{thebibliography}

\clearpage
\appendix

{
\section{TabUnroll Prompt Template} 
\label{sec:unrolled_tables}

You are a helpful AI assistant to help infer useful information from table structures. You are given a table in markdown format. Your goal is to write all the details conveyed in the table in the form of natural language statements. A statement is an atomic unit of information from the table.

Following the below instructions to do so:\\

\small
\begin{enumerate}
    \item Identify the column headers in the table.
    \item Identify the various rows in the table.
    \item From each row, identify meaningful and atomic pieces of information that cannot be broken down further.
    \item First, identify columns as primary key(s). A primary key is the column or columns that contain values that uniquely identify each row in a table.
    \item If there is only one primary key identified, use it and add information from each of the other columns one-by-one to form meaningful statements.
    \item If there are more than one primary key identified, use them and add information from each of the other columns one-by-one to form meaningful statements.
    \item If no primary key is detected, then form the statements by picking two columns at a time that make the most sense in a meaningful manner.
    \item In each of the above three cases, add information from other columns (beyond the primary key column(s) or the identified two columns in the absence of a primary key) only if it is necessary to differentiate repeating entities.
    \item Write all such statements in natural language.
    \item Do not exclude any detail that is present in the given table.
    \item Give the supporting rows for each atomic statement.
\end{enumerate}

Following are a few examples.

\subsection*{EXAMPLE 1}

\textbf{Title:} Koch

\noindent
\textbf{Table:}
\begin{Verbatim}[fontsize=\tiny]
|Year|     Competition     |         Venue        |Position|Event|Notes|
|----|---------------------|----------------------|--------|-----|-----|
|1966|European Indoor Games|Dortmund, West Germany|  1st   |400 m| 47.9| 
|1967|European Indoor Games|Prague, Czechoslovakia|  2nd   |400 m| 48.6| 
\end{Verbatim}
\noindent
\textbf{Statements:}
\begin{enumerate}[label=\arabic*.]
    \item European Indoor Games in 1966 occurred in Dortmund, West Germany.
    \item 1st position was obtained in the 1966 European Indoor Games.
    \item The 1966 European Indoor Games had a 400 m event.
    \item 47.9 in the 1966 European Indoor Games.
    \item European Indoor Games in 1967 occurred in Prague, Czechoslovakia.
    \item 2nd position was obtained in the 1967 European Indoor Games.
    \item The 1967 European Indoor Games had a 400 m event.
    \item 48.6 in the 1967 European Indoor Games.
\end{enumerate}
\noindent
\textbf{Rows:}
\begin{enumerate}[label=\arabic*.]
    \item | 1966 | European Indoor Games | Dortmund, West Germany | 1st | 400m | 47.9 |
    \item | 1967 | European Indoor Games | Prague, Czechoslovakia | 2nd | 400m | 48.6 |
\end{enumerate}
\noindent
\textbf{Example Bad Statements:}
\begin{enumerate}[label=\arabic*.]
    \item Koch came in 1st position in European Indoor Games in 1966 which occurred in Dortmund, West Germany.
    \item 47.9 in European Indoor Games in 1966 which occurred in Dortmund, West Germany.
    \item 2nd position in European Indoor Games in 1967 which occurred in Prague, Czechoslovakia.
\end{enumerate}

\subsection*{EXAMPLE 2}

\textbf{Title:} Isabella Rice - Film

\noindent
\textbf{Table:}
\begin{Verbatim}[fontsize=\tiny]
|Year|               Title                |        Role        |Notes|
|----|------------------------------------|--------------------|-----|
|2015|Kidnapped: The Hannah Anderson Story|   Becca McKinnon   | NaN | 
|2015|       Jem and the Holograms        |Young Jerrica Benton| NaN | 
|2015|             Asomatous              |    Sophie Gibbs    | NaN | 
|2017|           Unforgettable            |         Lily       | NaN | 
|2019|             Our Friend             |         Molly      | NaN |
\end{Verbatim}
\noindent
\textbf{Statements:}
\begin{enumerate}[label=\arabic*.]
    \item Kidnapped: The Hannah Anderson Story was filmed in 2015.
    \item Isabella Rice played the role of Becca McKinnon in Kidnapped: The Hannah Anderson Story.
    \item Jem and the Holograms was filmed in 2015.
    \item Isabella Rice played the role of Young Jerrica Benton in Jem and the Holograms.
    \item Asomatous was filmed in 2015.
    \item Isabella Rice played the role of Sophie Gibbs in Asomatous.
    \item Unforgettable was filmed in 2017.
    \item Isabella Rice played the role of Lily in Unforgettable.
    \item Our Friend was filmed in 2019.
    \item Isabella Rice played the role of Molly in Our Friend.
\end{enumerate}
\noindent
\textbf{Rows:}
\begin{enumerate}[label=\arabic*.]
    \item | 2015 | Kidnapped: The Hannah Anderson Story | Becca McKinnon | NaN |
    \item | 2015 | Jem and the Holograms | Young Jerrica Benton | NaN |
    \item | 2015 | Asomatous | Sophie Gibbs | NaN |
    \item | 2017 | Unforgettable | Lily | NaN |
    \item | 2019 | Our Friend | Molly | NaN |
\end{enumerate}
\noindent
\textbf{Example Bad Statements:}
\begin{enumerate}[label=\arabic*.]
    \item Isabella Rice played the role of Becca McKinnon in Kidnapped: The Hannah Anderson Story in 2015.
    \item Jem and the Holograms was filmed in 2015 where Isabella Rice played the role of Young Jerrica Benton.
    \item Isabella Rice played the role of Sophie Gibbs in Asomatous in 2015.
\end{enumerate}
}

\section{\datasetname Samples} 
\label{sec:dataset}

\subsection{Sample 1}

\textbf{Table Text}

Muarajati I, with a quay length of 275 meters and a depth of 7.0 meters at Low Water Springs (LWS), stands out as a robust terminal with a capacity of 3 tons per square meter. Muarajati II, featuring a quay length of 248 meters and a depth of 5.5 meters at LWS, offers a solid infrastructure with a capacity of 2 tons per square meter. Muarajati III, although more modest in size with an 80-meter quay length, matches Muarajati I in depth at 7.0 meters and a capacity of 3 tons per square meter. Linggarjati I, with a quay length of 131 meters and a depth of 4.5 meters at LWS, is a versatile berth with a capacity of 2 tons per square meter. Additionally, the port includes Pelita I, II, and III jetties, each featuring different lengths (30, 51, and 30 meters, respectively), all sharing a depth of 4.5 meters at LWS and a capacity of 1 ton per square meter.

\noindent
\textbf{Table Intent}

Principal cargo berths – Port of Cirebon

\noindent
\textbf{Table}

\begin{figure}[H]
    \centering
    \includegraphics[width=\columnwidth]{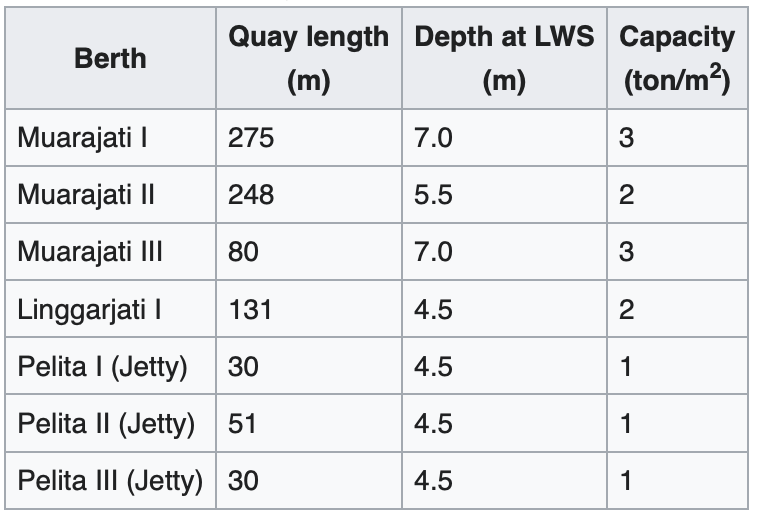}
\end{figure}

\subsection{Sample 2}

\textbf{Table Text}

In 2010, the television series "Glee" secured a nomination in the Choice Music: Group category. Four years later, in 2014, the animated film "Frozen" earned a nomination in the Choice Music: Single category, but it was in the category of Choice Animated Movie: Voice that the project achieved success, clinching the victory for its outstanding voice performance.

\noindent
\textbf{Table Intent}

Teen Choice Awards

\noindent
\textbf{Table}

\begin{figure}[H]
    \centering
    \includegraphics[width=\columnwidth]{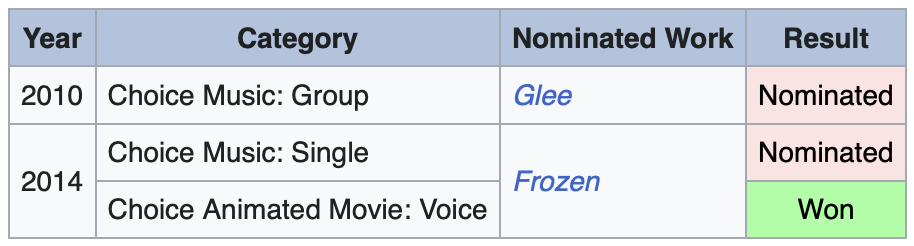}
\end{figure}

\subsection{Sample 3}

\textbf{Table Text}

Béranger Bosse, participating in the Men's 100m sprint, demonstrated impressive speed with a recorded time of 10.51 seconds during the heat, earning him a commendable 6th place. However, his journey concluded at the quarterfinal stage, as he fell short of advancing to the subsequent quarterfinal, semifinal and final rounds. Meanwhile, Mireille Derebona faced a setback in the Women's 800m, encountering disqualification in the heat. Consequently, there is no available data for her quarterfinal performance. Regrettably, Mireille did not progress to the later stages of the competition, missing out on the opportunities presented in the semifinal and final rounds.

\noindent
\textbf{Table Intent}

Athletic Performances of Béranger Bosse and Mireille Derebona in the 2008 Summer Olympics 

\noindent
\textbf{Table}

\begin{figure}[H]
    \centering
    \includegraphics[width=\columnwidth]{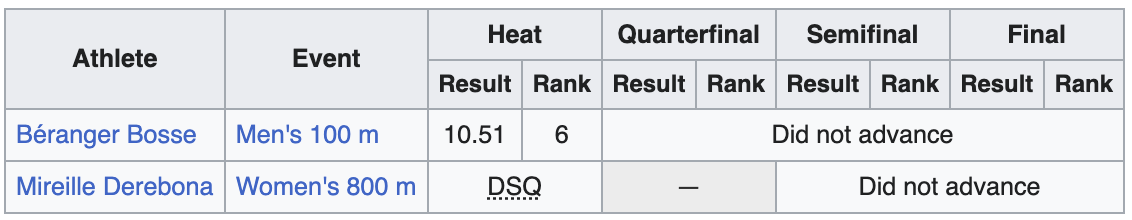}
\end{figure}

\section{Text-to-Table Prompt}
\label{sec:table_prompt}

\begin{verbatim}
Construct a table from a text. Ensure the 
column names are appropriate. Output in 
markdown format. Mark empty cells with 
"NaN".

Output only the final table.

EXAMPLES:
<FEW-SHOT EXAMPLES DEPENDING ON DATASET,
k=10>

TEXT:
{text}

TABLE:
\end{verbatim}

\section{Human Survey}
\label{sec:human}

\begin{figure}[ht]
    \centering
    \includegraphics[width=0.45\textwidth]{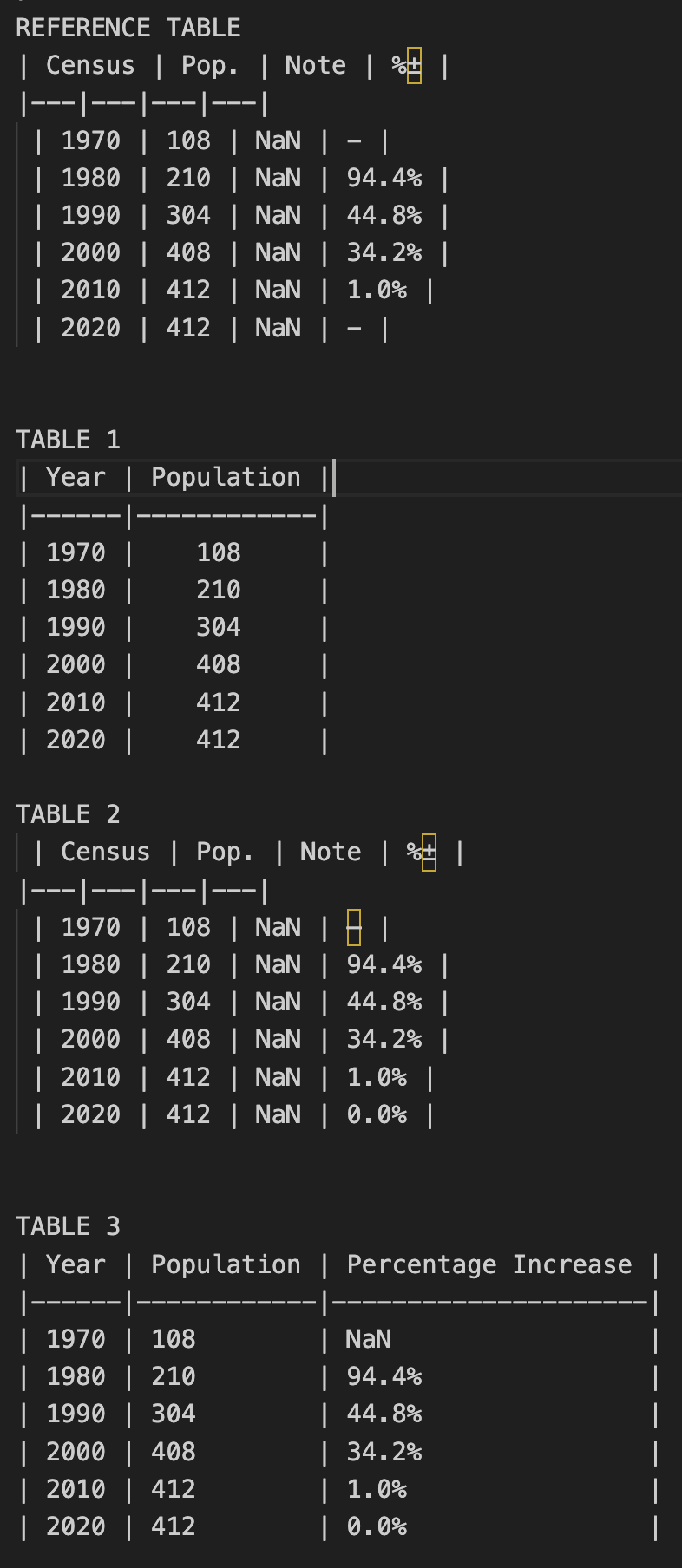} 
    \caption{Screenshot of file given to raters for evaluation.}
    \label{fig:human}
\end{figure}

\textbf{Task Description}: We need your assistance to evaluate the quality of generated tables from text.

\noindent
\textbf{Survey Format}: You will be given a text, reference table and 4 model generated tables. You will be presented with a series of questions designed to assess the overall quality, correctness and completeness of the generated tables against the reference table.

\noindent
\textbf{Question Types}: You will be asked to rate certain aspects of the tables on a scale of 1-5. Please follow the instructions carefully.

Rate the generated tables for the following aspects:
\begin{enumerate}
    \item Overall Quality: How easily can you understand the contents of the generated table and how does it compare against the ground truth table? (Scale 1-5)
    
    -- Contents refer to data within the cells and the column headers.
    
    Score 1 Nothing can be understood from the table and is of poor quality
    
    Score 2 Needs significant revisions to improve table quality (including the way content is placed, additions and/or omissions of information)
    
    Score 3 Needs small improvements
    
    Score 4 I can understand the current table but would like to see it better represented
    
    Score 5 Perfect Table

    \item Completeness: Does the generated table represent all the information present in the reference table? (Scale 1-5)
    
    -- Information refers to the facts and other relevant data the table depicts.
    
    -- Check if the information represented by the table is correct

    Score 1 No information from the reference table is in the table.
    
    Score 2 Some information from the reference table is present in the table (about 50\%)
    
    Score 3 Most information is present in the table (50-90\%)
    
    Score 4 Missing at most 1 fact from the text.
    
    Score 5 Perfect Table

    \item Correctness/Accuracy: Are only the relevant information from reference table present in the table and is the information present factually correct? (Scale 1-5)

    -- Ensure to understand the position of content in the table to determine if the correct facts are being conveyed.

    --Penalise the presence of unnecessary information in the table.

    --Infer what all information gets affected if one cell is incorrect.

    Score 1 Less than 10\% of the information is correct in the generated table.
    
    Score 2 Some unnecessary information and incorrect information is present in the table (greater than 30\% of table is unnecessary or incorrect)
    
    Score 3 Some unnecessary information is present in the table (less than 30\% of table is unnecessary or incorrect)
    
    Score 4 At most 1 additional fact is unnecessary or incorrect for the table.
    
    Score 5 Perfect Table

\end{enumerate}

\section{Human Validation of Unrolled Statements}
\begin{figure*}[ht]
    \centering
    \includegraphics[width=0.6\linewidth]{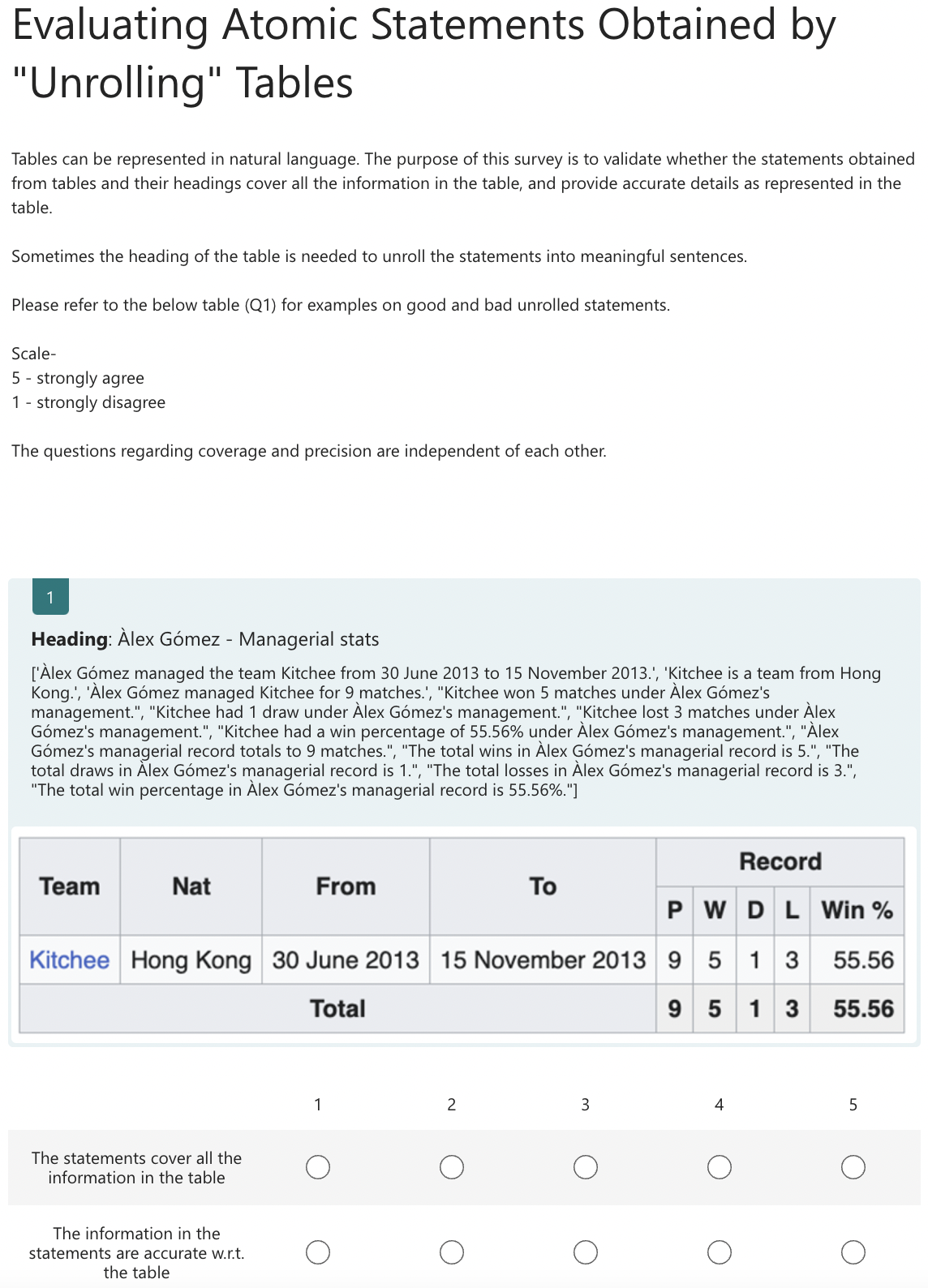} 
    \caption{Screenshot of Microsoft Forms used for survey.}
    \label{fig:survey}
\end{figure*}

\begin{figure*}[ht]
    \centering
    \includegraphics[width=0.6\linewidth]{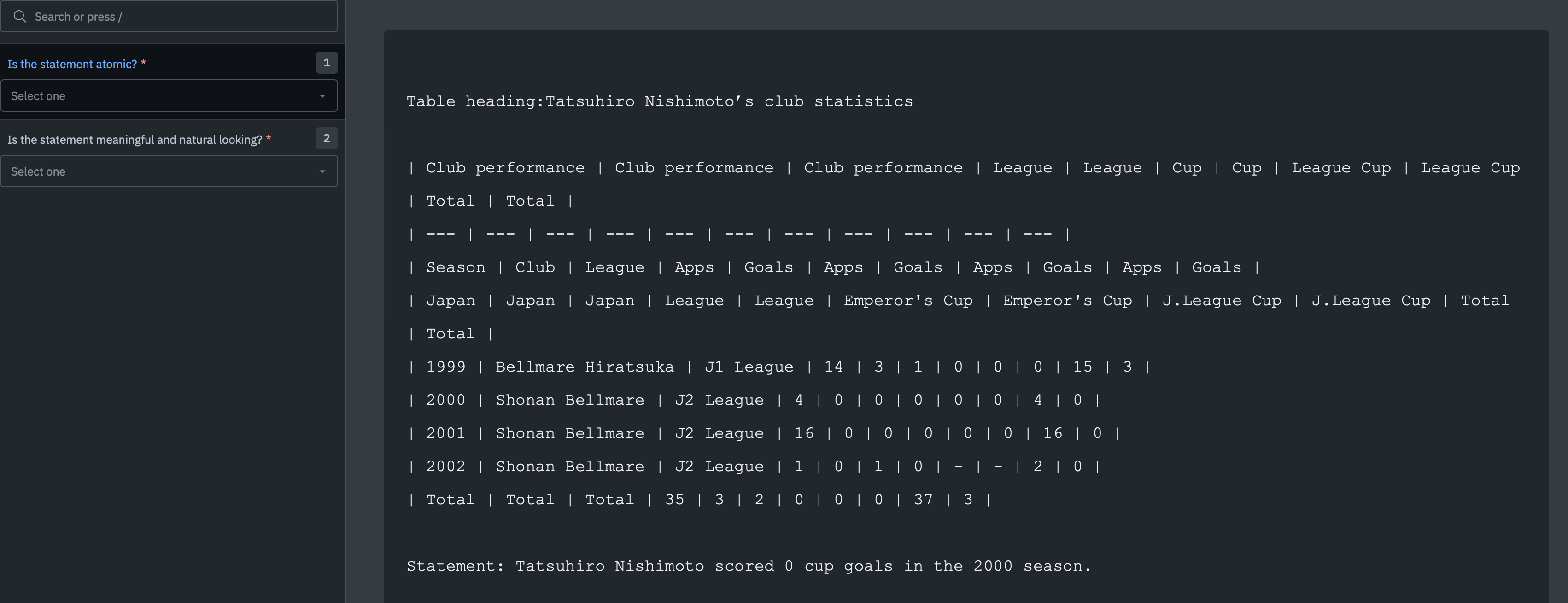} 
    \caption{Screenshot of the annotation for atomicity and meaningfulness.}
    \label{fig:survey_2}
\end{figure*}

Figures \ref{fig:survey} and \ref{fig:survey_2} illustrate the survey format for obtaining ratings for the quality of unrolled statements. Participants in the survey are asked to rate the unrolled statements based on:
\begin{enumerate}
    \item \textbf{Coverage}: Whether the statements encompass all the information provided in the table.
    \item \textbf{Precision}: The accuracy of the statements relative to the data in the table.
    \item \textbf{Atomicity}: If the statements can be broken down further into meaningful sentences by excluding information from specific columns.
    \item \textbf{Meaningfulness}: If the statements are meaningful and natural looking, based on the given table and intent.
\end{enumerate}
We hire three female annotators of Asian origin (from Philippines) for these surveys.
They are compensated at $\$10-15$ per hour.

\label{sec:valid}

\end{document}